\documentclass{article}

\usepackage{arxiv}

\usepackage[utf8]{inputenc} 
\usepackage[T1]{fontenc}    
\usepackage{hyperref}       
\usepackage{url}            
\usepackage{booktabs}       
\usepackage{amsfonts}       
\usepackage{nicefrac}       
\usepackage{microtype}      
\usepackage{graphicx}
\graphicspath{ {./images/} }
\usepackage[ruled,vlined]{algorithm2e}
\usepackage{siunitx}
\usepackage{textcomp}
\usepackage{float}
\usepackage{lscape}

\title{Deep Reinforcement Learning for Process Synthesis}

\author{
  Laurence I. Midgley \\
  Department of Engineering\\
  University of Cambridge\\
  Cambridge, United Kingdom \\
  and, \\
  Department of Chemical Engineering, \\
  University of Cape Town, \\
  Cape Town, South Africa \\
  \\
  \texttt{laurencemidgley@gmail.com} \\
}

\begin{document}
\maketitle

\begin{abstract}
This paper demonstrates the application of reinforcement learning (RL) to process synthesis by presenting Distillation Gym, a set of RL environments in which an RL agent is tasked with designing a distillation train, given a user defined multi-component feed stream. Distillation Gym interfaces with a process simulator (COCO and ChemSep) to simulate the environment. A demonstration of two distillation problem examples are discussed in this paper (a Benzene, Toluene, P-xylene separation problem and a hydrocarbon separation problem), in which a deep RL agent is successfully able to learn within Distillation Gym to produce reasonable designs. Finally, this paper proposes the creation of Chemical Engineering Gym, an all-purpose reinforcement learning software toolkit for chemical engineering process synthesis.  
\end{abstract}



\section{Introduction}
Reinforcement learning (RL), is a type of machine learning in which an agent makes a sequence of decisions within an environment to maximize an expected reward. RL has had many recent successful applications, including mastering games such as chess and Go \cite{silver_general_2018}.
Reinforcement learning has been recently applied to chemical engineering problems, notably process control \cite{nian_review_2020}. Computer aided process synthesis are currently dominated by optimisation techniques such as MINLP \cite{chen_recent_2017}. Preliminary work has been done showing the possibility of reinforcement learning for process synthesis using a simple problems simulated using a hand-crafted simulator \cite{midgley_reinforcement_2019}. This paper builds on this work to present a clear demonstration of RL for process synthesis.\footnote{Code available at https://github.com/lollcat/DistillationTrain-Gym}

\section{Reinforcement learning background}
\subsection{Reinforcement learning task definition}
\begin{figure}[h]
  \centering
    \includegraphics[scale=0.5]{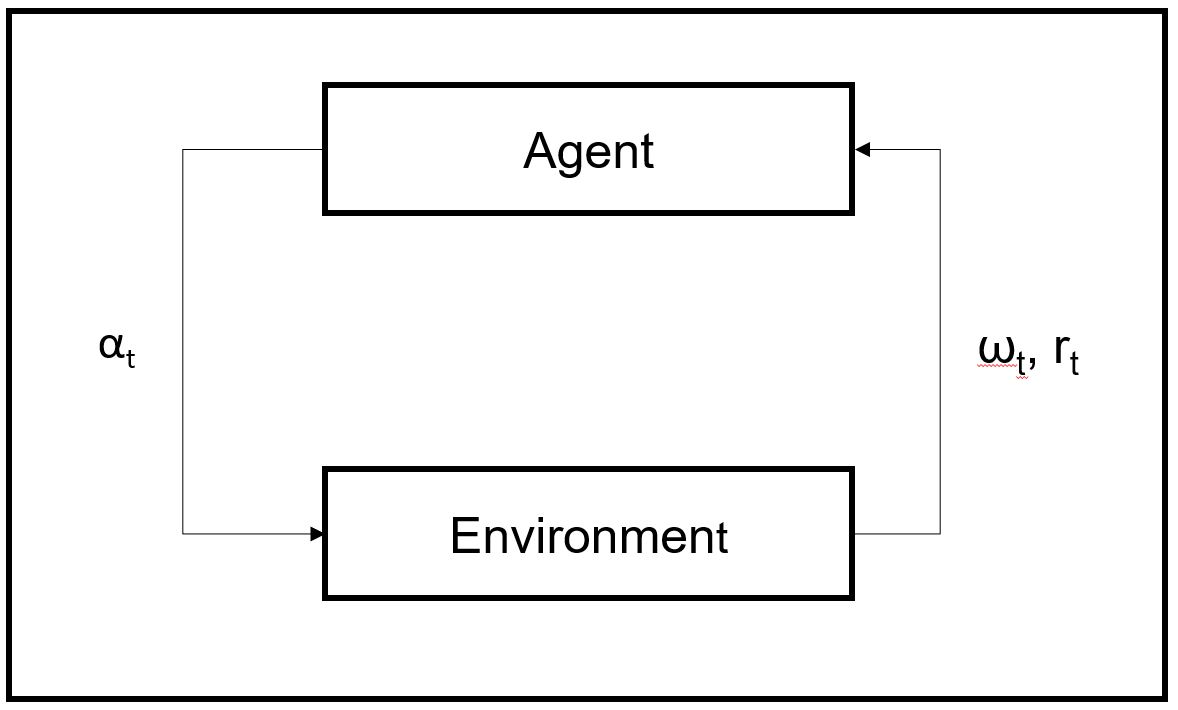}
    \caption{Markov Decision Process Diagram}
    \label{fig:MDP}
\end{figure}

Reinforcement learning tasks are structured as a Markov Decision Process (MDP)(Figure \ref{fig:MDP}). In an MDP, for each time step of the environment, the agent (1) makes an observation (or partial observation) $\omega_{t}$, of the state, of the environment $s_{t}$, (2) selects an action $\alpha_{t}$ from the set of possible actions A, based on the agent’s policy $\pi$, after which (3) the environment transitions to a new state $s_{t+1}$, and (4) the agent receives reward $r_{t}$. The transition of the environment may be stochastic or deterministic. The environment starts in a given state and follows this sequence of steps until the terminal state is reached. The goal of the agent is to learn a policy that maximizes the total expected reward it receives during the episode, given by $\max_{\pi}{E}\left[\sum_{t} r_t\right]$. 

\subsection{Soft Actor Critic}
This paper utilizes an adapted version of the Soft Actor Critic (SAC) agent \cite{haarnoja_soft_2018, haarnoja_soft_2018-1}.  This section gives a short summary of SAC, while Section \ref{sssec:Example_Overview} describes the adaption added to SAC to fit Distillation Gym. 
SAC utilizes an actor critic architecture where the Q-function, which predicts the value of a state conditional on an action, is approximated by a neural network (the “critic”), while the “actor” is a neural network that produces a set of actions aimed at maximizing the Q-value estimated by the “critic”. Instead of the typical value function, SAC uses a soft value function with an additional entropy regularization term that encourages exploration.  The soft Q-function is given by,

\begin{equation}
Q^{\pi}(s, a)={\mathrm{E}}_{s_{t} \sim P, \alpha' \sim \pi}[\sum_{t=0} \gamma^{t} R\left(s_{t}, a_{t}, s_{t+1}\right)+\alpha \sum_{t=1} \gamma^{t} H\left(\pi\left(s_{t}\right)\right) \mid s_{0}=s, a_{0}=a]
\end{equation}

where H, is the entropy of the policy at a given time step and $\alpha$ is a temperature parameter controlling the trade-off between maximising the reward and maximising the policy's entropy. The relation of the soft Q-function the soft value-function is then given by, 
\begin{equation}
\label{eq:soft-Q-V-relation}
V\left(\mathbf{s}_{t}\right)=\mathbb{E}_{\mathbf{a}_{t} \sim \pi}\left[Q\left(\mathbf{s}_{t}, \mathbf{a}_{t}\right)-\alpha \log \pi\left(\mathbf{a}_{t} \mid \mathbf{s}_{t}\right)\right]\end{equation}

The parameters $\theta$, of the critic network, can be found by minimising, 
\begin{equation}J_{Q}(\theta)=\mathbb{E}_{\left(\mathbf{s}_{t}, \mathbf{a}_{t}\right) \sim \mathcal{D}}\left[\frac{1}{2}\left(Q_{\theta}\left(\mathbf{s}_{t}, \mathbf{a}_{t}\right)-\left(r\left(\mathbf{s}_{t}, \mathbf{a}_{t}\right)+\gamma \mathbb{E}_{\mathbf{s}_{t+1} \sim p}\left[V_{\bar{\theta}}\left(\mathbf{s}_{t+1}\right)\right]\right)\right)^{2}\right]\end{equation}
Where $\mathcal{D}$ represents a replay buffer of experience. The term for the value-function is implicitly modelled through the relation to the Q-value given in Equation \ref{eq:soft-Q-V-relation}.  The parameters $\phi$, of the actor network can be found by minimizing, 
\begin{equation}
J_{\pi}(\phi)=\mathbb{E}_{\mathbf{s}_{t} \sim \mathcal{D}}\left[\mathbb{E}_{\mathbf{a}_{t} \sim \pi_{\phi}}\left[\alpha \log \left(\pi_{\phi}\left(\mathbf{a}_{t} \mid \mathbf{s}_{t}\right)\right)-Q_{\theta}\left(\mathbf{s}_{t}, \mathbf{s}_{t}\right)\right]\right]
\end{equation}
A “reparameterization trick” is then used where the policy reparametrized using, 
\begin{equation}
\mathbf{a}_{t}=f_{\phi}\left(\epsilon_{t} ; \mathbf{s}_{t}\right)
\end{equation}
where, $\epsilon_t$ is noise sample from an independent distribution (e.g. spherical Gaussian). Which allows for the actor cost function to be given as,
\begin{equation}
J_{\pi}(\phi)=\mathbb{E}_{\mathbf{s}_{t} \sim \mathcal{D}, \epsilon_{t} \sim \mathcal{N}}\left[\alpha \log \pi_{\phi}\left(f_{\phi}\left(\epsilon_{t} ; \mathbf{s}_{t}\right) \mid \mathbf{s}_{t}\right)-Q_{\theta}\left(\mathbf{s}_{t}, f_{\phi}\left(\epsilon_{t} ; \mathbf{s}_{t}\right)\right)\right]
\end{equation}

The temperature parameter, $\alpha$, can be set to a constant, or tuned throughout training for improved performance. In order to prevent over-estimation of the Q-function in the training of the actor, SAC uses two Q-networks, where the minimum Q-value of the networks is taken \cite{hasselt_double_2010, fujimoto_addressing_2018}. Target Q-networks with soft-updates are also used to stabilize training \cite{mnih_human-level_2015}. For a detailed description of SAC, see \cite{haarnoja_soft_2018, haarnoja_soft_2018-1}.

\section{Distillation Gym}
\subsection{Overall Description}
\begin{figure}[h]
  \centering
    \includegraphics[scale=0.5]{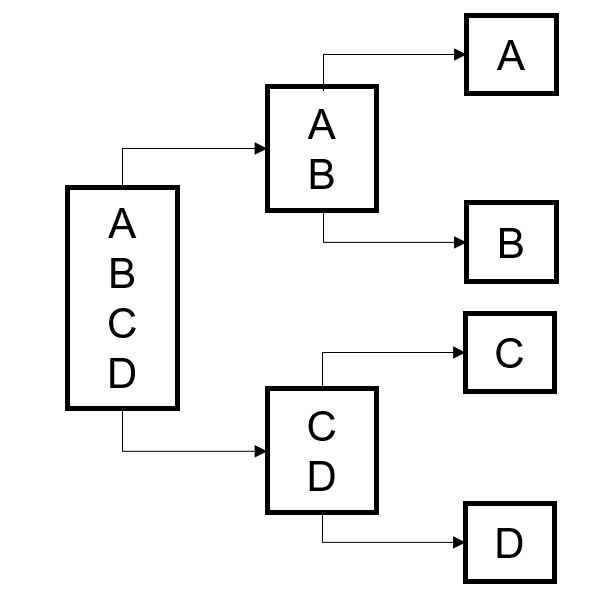}
    \caption{Distillation Gym's simple tree-structured column design}
    \label{fig:tree-MDP}
\end{figure}

Distillation Gym is a reinforcement learning environment involving the synthesis of simple distillation trains to separate compounds. In Distillation Gym, an RL agent is tasked with designing a common tree structured simple distillation train (Figure \ref{fig:tree-MDP}), which separates a user defined multicomponent stream. The user is tasked with instantiating a specific problem within distillation gym, defining: 
\begin{enumerate}
    \item The simulation system: component specification, physical and chemical properties of the system
    \item  Feed stream specification: component flowrates, conditions (temperature, pressure)
    \item Product definition: required purity, pure product selling stream prices
\end{enumerate}
Given the user specified problem, the RL agent is tasked with designing a distillation train in order to maximize the expected return. For a given step of the environment the reward is given,
\begin{equation}
    r(s_{t}, \alpha_{t}) = revenue - TAC
\end{equation}
where TAC is the total annual cost of the additional column specified by action $\alpha_{t}$, and revenue is the annual revenue corresponding to the sum of the value of any product produced in the distillate and bottoms of the additional column. The distillation train is designed through sequential addition of new columns to the existing process, where the agent specifies each column’s pressure (controlled by a valve before the column), number of stages, reflux ratio and reboil ratio. The agent operates only through the addition of new distillation columns to stream with unconnected endpoints, which results in the simple distillation train’s resulting simple tree structure. There are no other permitted changes to the process design (e.g.  adding recycle streams or editing existing units). All unit simulation is performed using COCO and ChemSep \cite{van_baten_coco_2020, kooijman_chemsep_2020}. The goal of Distillation Gym is to create a simple illustration of RL’s application to process synthesis. Interfacing with a process simulator, is a key component of this demonstration, as more general process synthesis RL tasks would most likely also follow this form.

\subsection{Markov Decision Process Framing}
Current most computational process synthesis techniques typically frame process synthesis as an optimisation problem where an objective function governed by a strict set of parameters and equations is maximised. Defining process synthesis as a reinforcement learning tasks, instead frames process synthesis as a sequential decision-making task (MDP). To define the MDP for process synthesis, the state, action, state-transition and reward structure need to be defined. The state should represent all the necessary information describing the current process configuration necessary for future design decisions. Typically for process synthesis tasks, the state should therefore most likely include information on all of the streams (resembling the stream table), as well as relevant additional information (e.g. on the unit operations).  To feed the state into a neural network, the array that describes the state needs to be of a fixed shape. For process synthesis, this constraint can cause some difficulty as the number of streams changes as the process is designed. One solution to this is to start with a large empty state (resembling a stream table populated with 0’s), which can be filled as the process is designed. The action should represent changes to the design of the process configuration (e.g. adding a new unit operation). The state transition is the running of the process simulation to calculate the updated information describing the process at its current point of design (e.g. calculating the flowrates in all of the streams). The reward describes the degree to which the objective of the design task (e.g. maximizing profit) is achieved. One intuitive way to think of process synthesis as an MDP is to imagine a human designing a process within a process simulator; what actions do they take? What information is relevant to their decisions? What is their goal?

In the Distillation Gym, because of the specific structure of the distillation problem, an adjusted form of the MDP can be created which allows for a significant simplification to the state. Specifically, instead of the state describing the process as a whole, through the change to the structure of the MDP, the state can instead describe a single stream. Algorithm \ref{env_pseudocode} below gives a description of the MDP for Distillation Gym. 

\begin{algorithm}[H]
\label{env_pseudocode}
\DontPrintSemicolon
\SetAlgoLined
initialization \tcp*{configure flowsheet}
D = deque() \tcp*{deque of the process streams with unconnected ends}
D.append(FeedStream)\;
done = False\;
\While{not done}{
   CurrentStream = D[0]\;
   observe CurrentStream.state\;
   choose whether to separate CurrentStream\;
   \eIf{chose to seperate}{
      specify column \tcp*{inlet pressure, number of stages, reboil ratio, reflux ratio}
      simulate flowsheet \tcp*{using interface with COCO simulator}
      reward = - TAC\;
      \For{Stream $\in$ (DistillateStream, BottomsStream)}{
         \eIf{Stream.purity >= PuritySpecification}{
            reward += Stream.revenue\;
            }{
            \tcp{if stream is doesn't meet product specification then it may be further separated}
            D.append(Stream)\; 
            }}
     \tcp{Can now remove current stream from D as it is now attached to a column}
      D.popleft()
      }{
   \tcp{if decide not to separate then CurrentStream becomes outlet to process so is removed from D}
     D.popleft() \;
    }
   \If{D is empty or max steps have been reached}{
      done = True \tcp*{episode is over}
      }}
\caption{Distillation Gym Environment Episode Structure}
\end{algorithm}

The reason this structure is possible is because actions only effect streams downstream in the process. This means that it is only necessary for the current stream being separated to be fed to the agent as the state, as the other process streams are completely independent to this decision. As shown in Algorithm \ref{env_pseudocode}, instead of standard linear MDP structure (state $\rightarrow$ next-state), a tree shaped decision process (referred to as tree-MDP) is used, where each state produces two next states (corresponding to the tops and bottoms of the distillation column). Instead of the next state coming immediately following the state, D, the deque of next states (unconnected streams) is sequentially stepped through. The episode is over when either (1) the maximum number of steps are reached or (2) all of the leaf node states are terminal, which is equivalent to D being empty. This tree structured decision process can also be interpreted as a partially observable MDP where the agent observes the current stream that it is separating, and both the outlet streams that the separation produces instead of observing the whole system. Total episode reward can still be calculated by summing the rewards over all time steps. However, the value of a state is now given by the immediate reward, and the value of all of the future rewards of states downstream in the tree decision process structure - as shown by the recursively defined value function below. 

\begin{equation}V(s)_{\pi}=E_{s \prime \sim P, a \sim \pi}\left[r+\gamma V\left(s_{\text {tops}}^{\prime}\right)_{\pi}+\gamma V\left(s_{\text {bottoms}}^{\prime}\right)_{\pi}\right]
\end{equation}

\subsection{Distillation train synthesis examples}

\subsubsection{Overview}
\label{sssec:Example_Overview}

Two example problems are given below, both largely based (in terms of feed composition and property package definitions) on examples from the ChemSep example page \cite{kooijman_chemsep_2020} and are common distillation problems. In both problems a Soft Actor Critic agent was adapted to include the tree-MDP, using the adapted soft Q-function, as shown in the recursively defined Q-function below.
\begin{equation}
\label{eqn:tree-MDP-softQ}
Q(s, a)_{\pi}=E_{s^{\prime} \sim P, a, \sim \pi}\left[r+\gamma\left(Q\left(s_{\text {tops}}^{\prime}, a_{\text {tops}}^{\prime}\right)-\alpha H\left(\pi\left(s_{\text {tops}}^{\prime}\right)\right)\right.\right.
+\gamma\left(Q\left(s_{\text {bottoms}}^{\prime}, a_{\text {bottoms}}^{\prime}\right)-\alpha H\left(\pi\left(s_{\text {bottoms}}^{\prime}\right)\right)\right]
\end{equation}
The agents action of whether or not to separate a specific stream is determined by whether the soft Q-value was positive (choose separate) or negative (choose not to separate). A consequence of this is that the Q-value of the next state (used in Equation \ref{eqn:tree-MDP-softQ}), has a lower bound of zero - because in situations where the Q-value was negative, the agent would decide not to separate the stream, causing the state to be a terminal leaf. Because the remainder of the actions (column specification) are continuous, this addition allowed for a simple extension to add the binary decision of deciding whether or not to separate a given stream. 

\begin{figure}[H]
  \centering
    \includegraphics[scale=1]{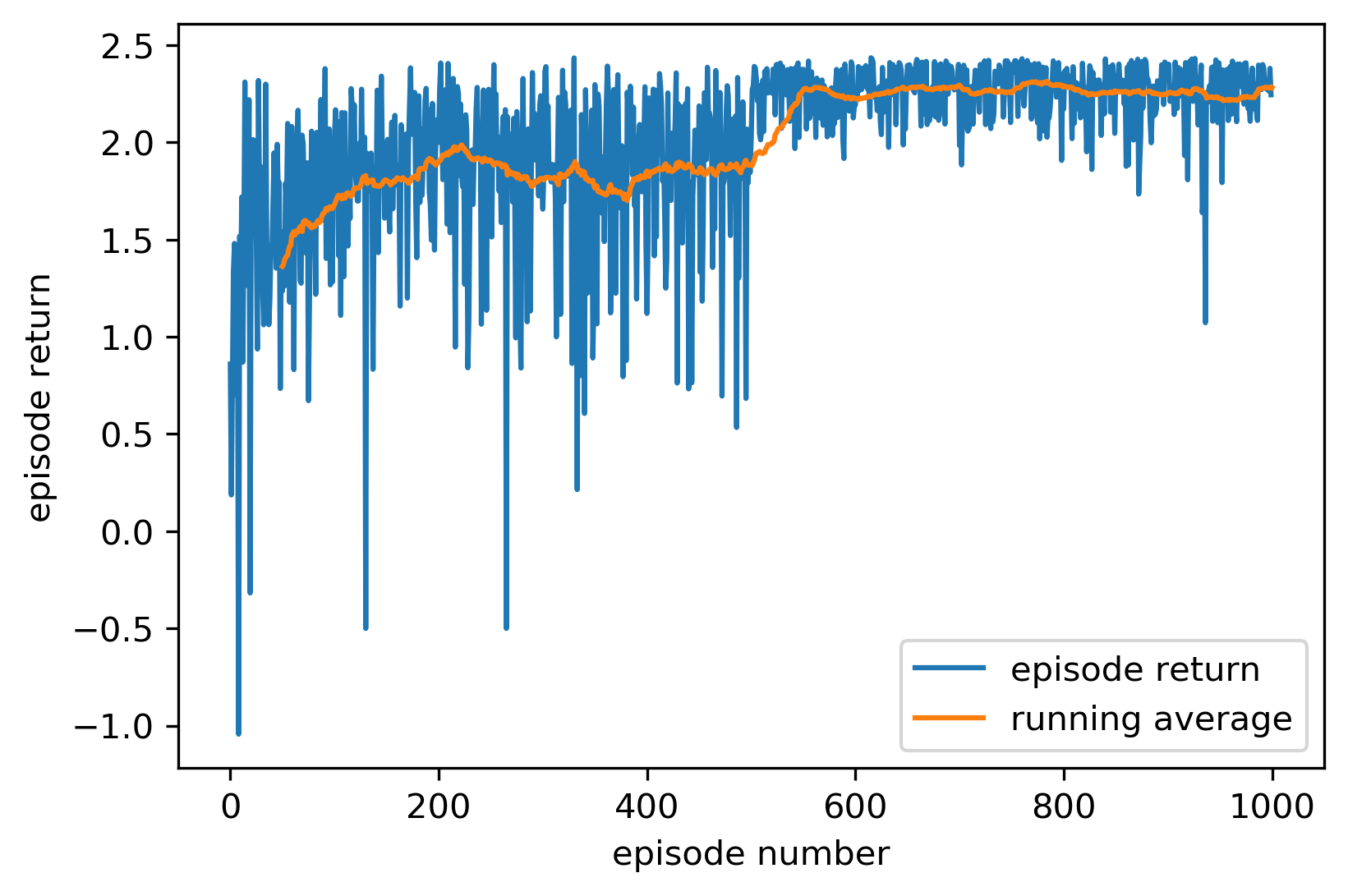}
    \caption{Agent training on BTX problem}
    \label{fig:BTX train}
\end{figure}

\begin{figure}[H]
  \centering
    \includegraphics[scale=1]{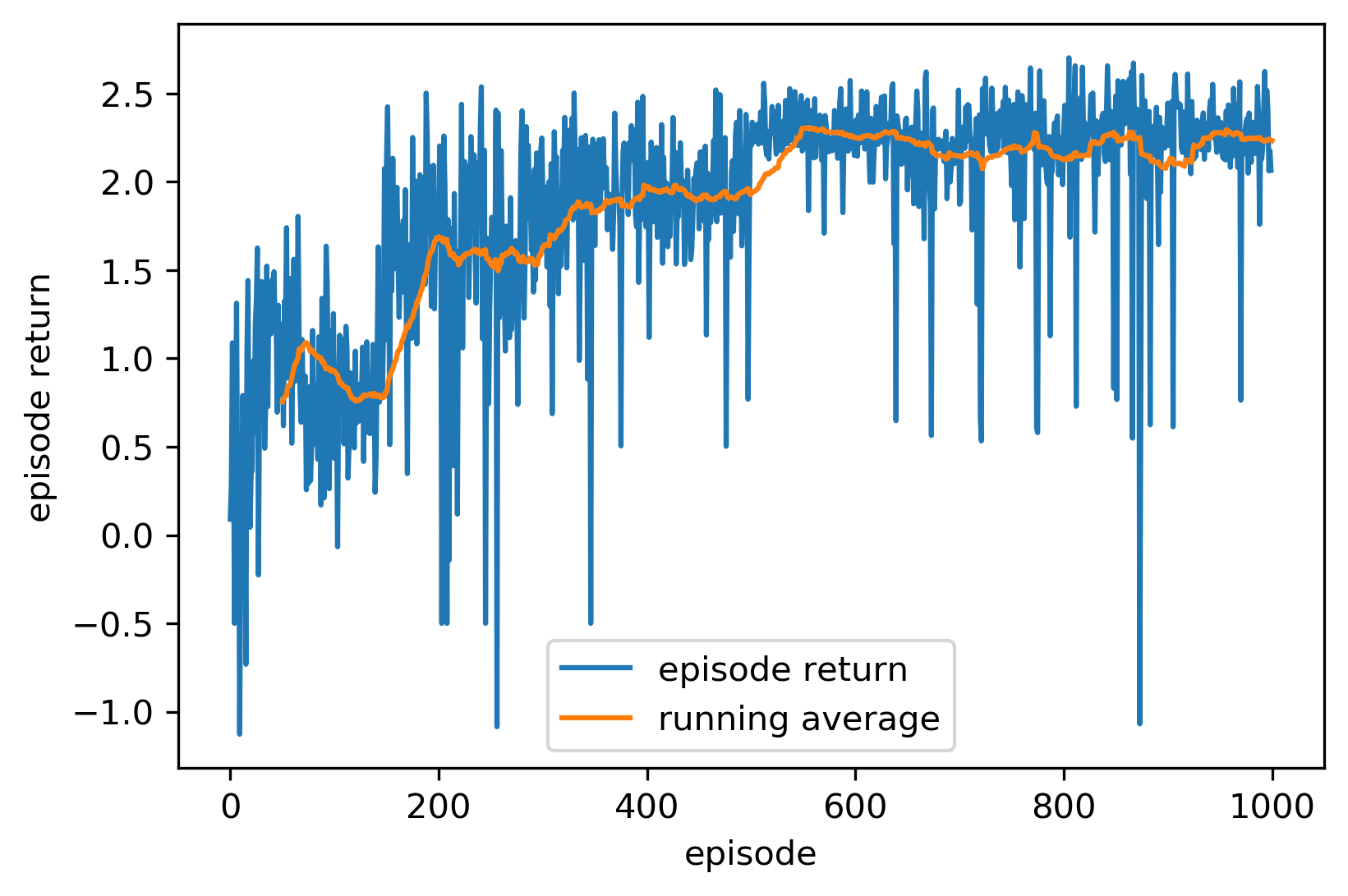}
    \caption{Agent training on Hydrocarbon problem}
    \label{fig:Hydrocarbon train}
\end{figure}

In both examples the agent was able to learn within the environment, to produce better and better distillation train configurations. In Figure \ref{fig:BTX train} and Figure \ref{fig:Hydrocarbon train}, this is shown by both the improvement in average return, and the improvement of the “best designs” (shown by the peaks in the return). In both examples, the agent is able to achieve progressively higher sum total revenue (through obtaining higher product recovery) and lower sum total TAC. For each example, autogenerated block-flow-diagrams (BFD) of the best design throughout training are given, as well as the key outcome metrics. Overall, these problems successfully demonstrate that the framing of process synthesis as a reinforcement learning task. 

\subsubsection{Problem 1: BTX separation}

\subsubsubsection{User specification of problem}
\begin{itemize}
    \item \textbf{Simulation System}: Benzene, Toluene and p-Xylene. Flowsheet settings copied from COCO file on ChemSep example page \cite{kooijman_chemsep_2020}
    \item \textbf{Starting stream definition}: Equimolar (3.35 mol/s of each compound) feed at 25\textdegree C, 1 atm
    \item \textbf{Product definition}: Required purity of 95\%, price (\$/tonne): 488, 488, 510 \cite{noauthor_echemi_2020}
\end{itemize}

\newpage
\subsubsubsection{Results}

\begin{table}[H]
 \caption{Performance metrics for best design outcome on BTX problem}
  \centering
  \begin{tabular}{ll}
  \toprule
Total Revenue & 	\$ 13.17 million \\
Total Column TAC &	\$ 0.45 million \\
    \bottomrule
  \end{tabular}
\end{table}

\begin{table}[H]
 \caption{Recoveries for best design outcome on BTX problem}
  \centering
  \begin{tabular}{ll}
    \toprule
    Compound  &  Recovery \\
    \midrule
    Benzene   &  98.2\% \\
    Toluene   &  99.7\% \\
    p-Xylene  &  > 99.9\% \\
    \bottomrule
  \end{tabular}
\end{table}

\begin{figure}[H]
  \centering
    \includegraphics[scale=0.7]{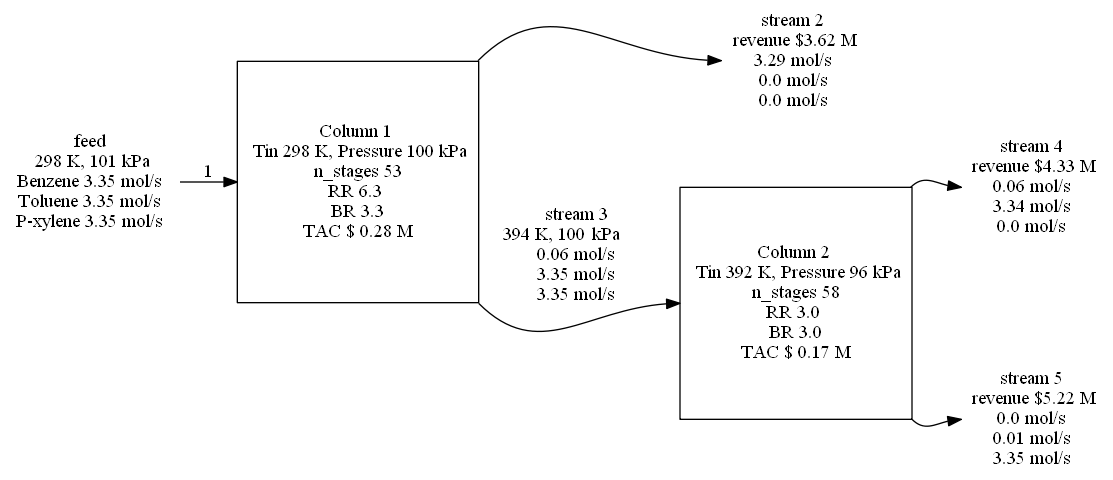}
    \caption{Best Design for BTX problem (autogenerated BFD)}
    \label{fig:BTX BFD}
    \small\textsuperscript{For simplicity, the autogenerated BFD has a single block for the valve-distillation column pair} 
\end{figure}

\subsubsection{Problem 2: Hydrocarbon separation}
\subsubsubsection{User specification of problem}
\begin{itemize}
    \item \textbf{Simulation System}: Ethane, Propane, Isobutane, N-butane, Isopentane, N-pentane. Flowsheet settings copied from COCO file on ChemSep example page \cite{kooijman_chemsep_2020}.
    \item \textbf{Starting stream definition} : Compound flows (mol/s): 17, 1110, 1198, 516, 334, 173. Conditions: 105{\textdegree}C, 17.4 atm. 
    \item \textbf{Product definition}: Required purity of 95\%, price (\$/tonne): 125, 204, 272, 249, 545, 545 \cite{eia_prices_2018}
\end{itemize}

\newpage
\subsubsubsection{Results}

\begin{table}[H]
 \caption{Performance metrics for best design outcome on Hydrocarbon problem}
  \centering
  \begin{tabular}{ll}
  \toprule
Total Revenue & 	\$ 1588  million \\
Total Column TAC &	\$ 119  million \\
    \bottomrule
  \end{tabular}
\end{table}

\begin{table}[H]
 \caption{Recoveries for best design outcome on Hydrocarbon problem}
  \centering
  \begin{tabular}{ll}
    \toprule
    Compound  &  Recovery \\
    \midrule
    Ethane &	0\% \\
    Propane	& 98.9\% \\
    Isobutane &	97.3\% \\
    N-butane &	91.1\% \\
    Isopentane &	99.6\% \\
    N-pentane &	97.0\% \\

    \bottomrule
  \end{tabular}
\end{table}

\begin{landscape}
\begin{figure}[H]
  \centering
    \includegraphics[scale=0.35]{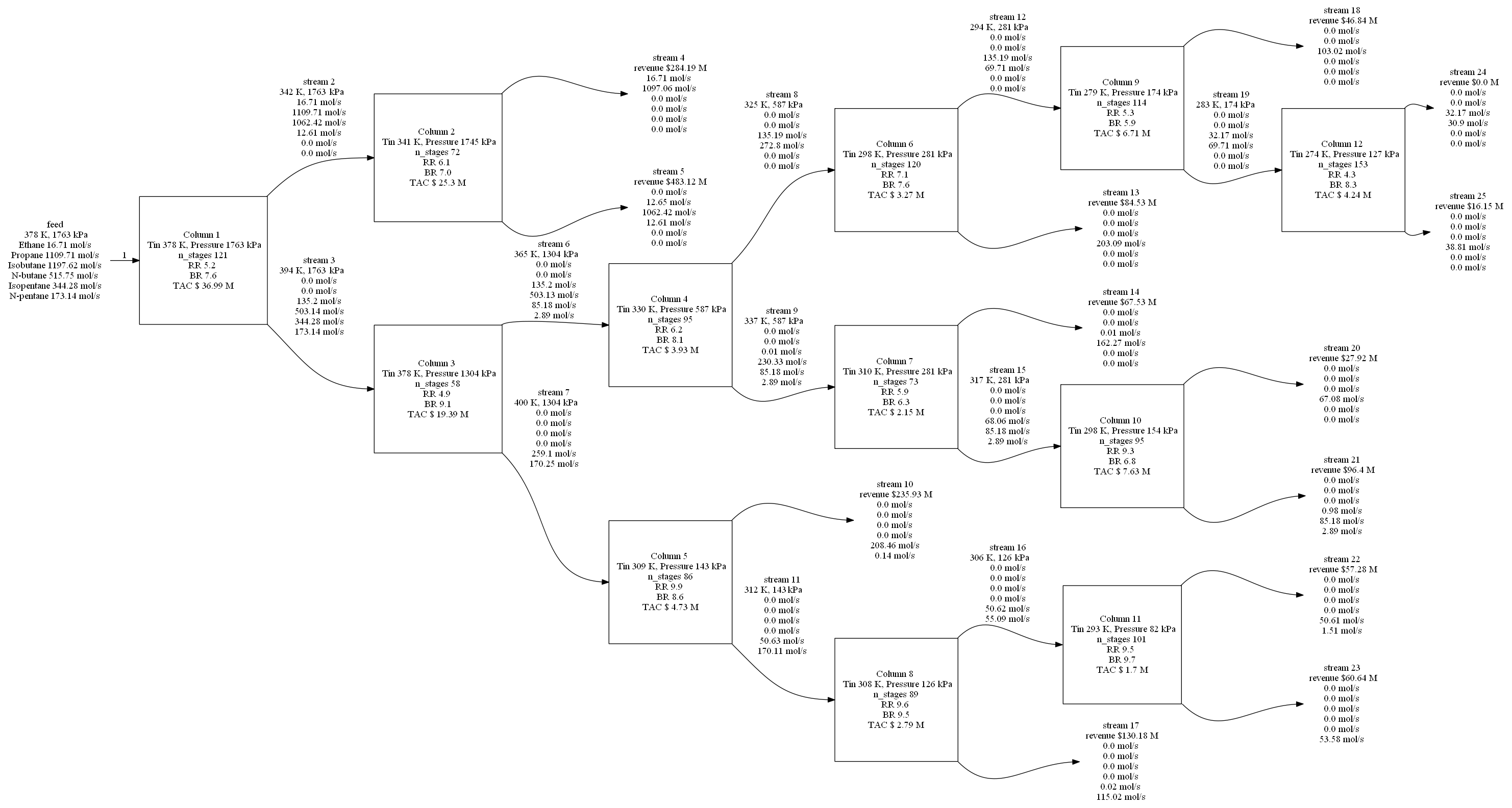}
    \caption{Best Design for Hydrocarbon problem}
    \label{fig:Hydrocarbon BFD}
\end{figure}
\end{landscape}

\section{Further developments \& Concluding Remarks}
Distillation Gym and the corresponding examples have purposely been designed to be simple process synthesis tasks, as the purpose of Distillation Gym is primarily as a demonstration of the possibility of RL for process synthesis. The key functions that Distillation Gym  demonstrates are (1) structuring process synthesis as a RL task (2) application of a RL agent, (3) interface with a process simulator (COCO simulator), (4) generality through user specified problem definition, (5) reasonable task complexity (branching process configuration, multivariate unit specification). 

The next step in the continuation of this project; Chemical Engineering Gym, an all-purpose software toolkit for generating process synthesis problems, structured as reinforcement learning tasks. Extending the RL for process synthesis approach to tasks is where RL may be advantageous relative to other conventional computational approaches, due to the ability of RL to solve more open-ended problems. More fundamentally, this paper proposes that the framing level of “process synthesis as a sequence of decisions” is more apt for general process synthesis problems than the framing as “optimizing a set of process parameters/equations”, which conventional optimization techniques follow. 

The key improvement that Chemical Engineering Gym would add, would be to allow problem specifications with a far larger action space. The problem specification should be able to include all of the relevant actions that a human process designer takes within a simulator’s GUI when designing a process. This would include allowing the agent to add a greater variety of unit operations to a process (e.g. reactors, heat exchangers), allowing the agent to add streams to the process (e.g. selecting when to “buy” raw materials, specified by a price dataset), allowing the agent greater ability to manipulate flowsheet topology (e.g. adding recycle streams), allowing the agent to select which streams exit the process (aided with a dataset of product prices/waste removal prices), editing existing unit parameters etc. In the extreme, with a large database of chemical prices, a reinforcement learning agent could be simply tasked with searching for novel profitable process designs, starting from a blank flowsheet, within the bounds of what can be accurately simulated. 

Chemical Engineering Gym would require a commercial process simulator(s) to interface with, in order to run the process simulation. Currently there is a large room for process simulators to improve interface with external programs (e.g. python). Process simulators do have some interface functionality with external programs, specifically they provide the ability to edit existing unit parameters, run the process simulation and retrieve results. However, much of the functionality in process simulator GUI is left out, most notably changing the flowsheet topology (e.g. through adding new units), which is a key component of process synthesis. Adding improvements to the interface between process simulators and eternal programs would most likely have benefits to the field of computational process synthesis in general. 

Progress in within the field of reinforcement learning has been significantly aided by freely available general purpose RL toolkits/frameworks, like OpenAI Gym \cite{brockman_openai_2016}. Similarly, creating a toolkit for RL within a process synthesis context could greatly aid research within the field. Distillation Gym presents the key fist step towards such a toolkit. 

\newpage
\bibliographystyle{unsrt}  %
\bibliography{references} 

\begin{thebibliography}{10}

\bibitem{silver_general_2018}
David Silver, Thomas Hubert, Julian Schrittwieser, Ioannis Antonoglou, Matthew
  Lai, Arthur Guez, Marc Lanctot, Laurent Sifre, Dharshan Kumaran, and Thore
  Graepel.
\newblock A general reinforcement learning algorithm that masters chess, shogi,
  and {Go} through self-play.
\newblock {\em Science}, 362(6419):1140--1144, 2018.
\newblock Publisher: American Association for the Advancement of Science.

\bibitem{nian_review_2020}
Rui Nian, Jinfeng Liu, and Biao Huang.
\newblock A review {On} reinforcement learning: {Introduction} and applications
  in industrial process control.
\newblock {\em Computers \& Chemical Engineering}, 139:106886, August 2020.

\bibitem{chen_recent_2017}
Qi~Chen and I.E. Grossmann.
\newblock Recent {Developments} and {Challenges} in {Optimization}-{Based}
  {Process} {Synthesis}.
\newblock {\em Annual Review of Chemical and Biomolecular Engineering},
  8(1):249--283, 2017.
\newblock \_eprint: https://doi.org/10.1146/annurev-chembioeng-080615-033546.

\bibitem{midgley_reinforcement_2019}
Laurence Midgley and Michael Thomson.
\newblock Reinforcement learning for chemical engineering process synthesis.
\newblock Technical report, Zenodo, November 2019.
\newblock https://zenodo.org/record/3556549\#.X2mx1mhKhPY.

\bibitem{haarnoja_soft_2018}
Tuomas Haarnoja, Aurick Zhou, Pieter Abbeel, and Sergey Levine.
\newblock Soft {Actor}-{Critic}: {Off}-{Policy} {Maximum} {Entropy} {Deep}
  {Reinforcement} {Learning} with a {Stochastic} {Actor}.
\newblock {\em arXiv:1801.01290 [cs, stat]}, August 2018.
\newblock arXiv: 1801.01290.

\bibitem{haarnoja_soft_2018-1}
Tuomas Haarnoja, Aurick Zhou, Kristian Hartikainen, George Tucker, Sehoon Ha,
  Jie Tan, Vikash Kumar, Henry Zhu, Abhishek Gupta, and Pieter Abbeel.
\newblock Soft actor-critic algorithms and applications.
\newblock {\em arXiv preprint arXiv:1812.05905}, 2018.

\bibitem{hasselt_double_2010}
Hado~V. Hasselt.
\newblock Double {Q}-learning.
\newblock In {\em Advances in neural information processing systems}, pages
  2613--2621, 2010.

\bibitem{fujimoto_addressing_2018}
Scott Fujimoto, Herke Van~Hoof, and David Meger.
\newblock Addressing function approximation error in actor-critic methods.
\newblock {\em arXiv preprint arXiv:1802.09477}, 2018.

\bibitem{mnih_human-level_2015}
Volodymyr Mnih, Koray Kavukcuoglu, David Silver, Andrei~A Rusu, Joel Veness,
  Marc~G Bellemare, Alex Graves, Martin Riedmiller, Andreas~K Fidjeland, and
  Georg Ostrovski.
\newblock Human-level control through deep reinforcement learning.
\newblock {\em Nature}, 518(7540):529, 2015.

\bibitem{van_baten_coco_2020}
Jasper van Baten and Richard Baur.
\newblock {COCO} - the {CAPE}-{OPEN} to {CAPE}-{OPEN} simulator, 2020.
\newblock https://www.cocosimulator.org/.

\bibitem{kooijman_chemsep_2020}
Harry Kooijman and Ross Taylor.
\newblock Chemsep, 2020.
\newblock http://www.chemsep.org/index.html.

\bibitem{noauthor_echemi_2020}
Echemi: {Provide} {Chemical} {Products} and {Services} to {Global} {Buyers},
  2020.
\newblock https://www.echemi.com.

\bibitem{eia_prices_2018}
EIA.
\newblock Prices for hydrocarbon gas liquids - {U}.{S}. {Energy} {Information}
  {Administration} ({EIA}), 2018.

\bibitem{brockman_openai_2016}
Greg Brockman, Vicki Cheung, Ludwig Pettersson, Jonas Schneider, John Schulman,
  Jie Tang, and Wojciech Zaremba.
\newblock {OpenAI} {Gym}, 2016.

\end{thebibliography}
\end{document}